\documentclass{article}

\usepackage[preprint]{icml2026}
\usepackage{natbib}
\usepackage{booktabs}
\usepackage{amsmath,amssymb}
\usepackage{graphicx}
\usepackage{microtype}
\usepackage{xspace}
\usepackage{enumitem}
\usepackage{url}
\usepackage[skins,breakable]{tcolorbox}

\definecolor{promptboxbg}{RGB}{248,250,252}
\definecolor{promptboxline}{RGB}{76,96,122}
\definecolor{promptboxtitle}{RGB}{54,73,96}
\newtcolorbox{promptbox}[1]{%
enhanced, breakable,
colback=promptboxbg, colframe=promptboxline,
colbacktitle=promptboxtitle, coltitle=white,
fonttitle=\bfseries\small, title={Prompt: #1},
boxrule=0.5pt, arc=1mm,
left=6pt,right=6pt,top=5pt,bottom=5pt,
before upper={\raggedright\parindent=0pt},
before skip=5pt, after skip=7pt
}
\newcommand{\promptfield}[2]{\noindent\textbf{#1.} #2\par\vspace{2pt}}
\newcommand{\promptcontext}[4]{\noindent\hangindent=1.25em\textbf{#1.} \texttt{weight=#2; node=#3;} text=\emph{#4}\par\vspace{1pt}}
\icmltitlerunning{HawkesLLM: Semantic Uncertainty Propagation}

\newcommand{\HawkLLM}{HawkesLLM\xspace}

\begin{document}

\twocolumn[
\icmltitle{HawkesLLM: Semantic Uncertainty Propagation in Agentic Text Simulation}

\begin{icmlauthorlist}
\icmlauthor{Zewei Deng}{umn}
\icmlauthor{Tinghan Ye}{gatech}
\icmlauthor{Liyan Xie}{umn}
\end{icmlauthorlist}
\icmlaffiliation{umn}{Department of Industrial and Systems Engineering, University of Minnesota}
\icmlaffiliation{gatech}{H. Milton Stewart School of Industrial and Systems Engineering, Georgia Institute of Technology}
\icmlcorrespondingauthor{Zewei Deng}{deng0379@umn.edu}

\vskip 0.22in
]
\printAffiliationsAndNotice{%
Accepted at the ICML 2026 Workshop on Statistical Frameworks for Uncertainty in Agentic Systems.
}

\begin{abstract}
Agentic text-simulation systems write in sequence, with each item becoming possible context for later steps. That makes uncertainty path-dependent: an early ambiguity can affect later outputs. This paper studies this problem with \HawkLLM, a framework that separates temporal influence modeling from text generation. We represent the cascade as a network whose nodes are text-generating agents. A multivariate Hawkes process models how these nodes activate over time and which earlier node outputs should influence later prompts. A language model then writes each new event from the compact memory selected by this temporal model. We evaluate the framework on a held-out Global Database of Events, Language, and Tone (GDELT) news-cascade case study. The diagnostics track semantic alignment with local held-out references and separate local drift from global drift. In this setting, \HawkLLM improves late-stage semantic alignment under a compact prompt-memory budget.
\end{abstract}

\section{Introduction}

Many text-simulation workflows unfold as multi-step processes. A system writes an item, keeps it in memory, and later writes with that memory in view. Here, \emph{agentic} refers to this iterative use of generated memory: each output can become part of the next prompt. In such systems, uncertainty belongs to the trajectory, not just to one response. Ambiguity, drift, and mismatch can carry forward because later prompts depend on earlier generated content.

Recent work on uncertainty in language-agent systems and semantic uncertainty in free-form generation motivates this view \citep{zhao2025uncertainty,han2024uncertaintyaware,liu2024toolcalibration,farquhar2024semanticentropy,lin2023generating}. If generated outputs become future context, uncertainty should be tracked across the sequence rather than only at the end.

We study this problem as \emph{semantic uncertainty propagation}: how generated events stay, or fail to stay, near local held-out reference neighborhoods as the generated history grows. The references serve as semantic anchors from the same event stream. They let us ask whether the simulation remains in the same local region over time.

We propose \HawkLLM for this setting. The simulation is a cascade over text-generating nodes. A Hawkes process models when nodes activate and which earlier node outputs should matter for the next prompt. The language model then writes the next event from this selected memory. We use \emph{memory selection} for the choice of node-specific predecessor texts and weights. This design is also motivated by evidence that long-context language models do not always use all provided context reliably \citep{liu2024lost}.

We demonstrate the framework on a GDELT news-cascade case study. The news domain is useful here because it has timestamped event streams that can be mapped to a small set of node categories. Held-out articles from the same topic window serve as imperfect local semantic references for generated events.

We make three contributions. First, we formulate semantic uncertainty propagation for iterative text simulation. Second, we develop \HawkLLM as a Hawkes-guided transition that samples event times and nodes and constructs compact node-wise prompt memory. Third, we evaluate it on the held-out news-cascade case study, where \HawkLLM improves late-stage alignment under a compact prompt-memory budget.

\section{Related Work}
One line of work asks how uncertainty should be handled inside multi-step agent systems \citep{zhao2025uncertainty,han2024uncertaintyaware,liu2024toolcalibration}. In that setting, uncertainty often becomes a control signal: the agent may call a tool, ask for help, or calibrate its answer. Here the object is the generated text path. Earlier generations become part of the state that later generations read, so we track how semantic uncertainty moves along the realized cascade.

Semantic-uncertainty and black-box confidence methods usually compare possible answers to the same free-form generation problem \citep{farquhar2024semanticentropy,lin2023generating}. Drift work looks at what happens after generation is repeated for many steps \citep{spataru2024know,mohamed2025broken}. We take the path itself as the object of evaluation: each generated event is compared with local held-out reporting, while the drift metrics separate short-range agreement with prompt memory from longer-range movement away from the seed.

Graph-cascade models give us a fixed graph, seed events, and a cascade unfolding over that graph \citep{kempe2015maximizing,kleinberg2007cascading}. In our setting, each activation also produces text. We use a Hawkes process for the temporal layer because it turns past activations into directed, decaying influence scores between nodes \citep{hawkes1971spectra,ogata1981lewis,rizoiu2017tutorial}. Neural temporal point processes and recent temporal-language hybrids can model richer dynamics or next-event prediction \citep{mei2017neural,zuo2020transformer,li2025rhythm,liu2025tppllm,zhou2025advances}. The simpler parametric Hawkes model is useful here because its fitted node-to-node influence can be exposed as an inspectable memory signal for the LLM prompt. In that sense, \HawkLLM sits between LLM-based social or agent simulation \citep{park2023generative,sun2024decoding,murdock2023reddit,zhang2025llmaidsim} and retrieval-augmented generation \citep{lewis2020retrieval}: the state is a sequence of generated event texts, and the retrieved memory is driven by temporal cascade dynamics.

\section{Methods}

\HawkLLM has two layers. The first is a text-simulation loop: a node is scheduled, a compact memory is selected, and the LLM writes the next event. The second is a Hawkes process that supplies the schedule and memory weights. Section~\ref{sec:framework} defines the generic loop. Section~\ref{sec:hawkes-process} introduces the multivariate Hawkes model for temporal influence among nodes. Section~\ref{sec:hawkes-propagation} then plugs that model into the loop.

\subsection{Agentic Text Simulation Framework}
\label{sec:framework}
We model text propagation on a fixed directed graph \citep{kempe2015maximizing,kleinberg2007cascading}. A node is a text-generating agent in the simulation, and an event is a timestamped activation of one node together with the text generated at that activation. Let \(\mathcal{G}_0=(\mathcal{N},\mathcal{E})\) denote the unweighted influence graph. Here \(\mathcal{N}=\{1,\ldots,N\}\) is the node set, and \(\mathcal{E}\subseteq\mathcal{N}\times\mathcal{N}\) is the set of allowed directed influences. We assume \(\mathcal{E}=\mathcal{N}\times\mathcal{N}\), so every node can influence every other node, including itself. Any node may be activated at a generation step, and any earlier activated node can contribute memory when allowed by \(\mathcal{E}\). Section~\ref{sec:hawkes-process} describes how we learn the effective edge weights from data.

Generation starts from a seed event \(e_0=(\tau_0,n_0,x_0)\), where \(\tau_0\) is the seed timestamp, \(n_0\in\mathcal{N}\) is the seed node, and \(x_0\) is the seed text. The realized cascade is the event history that grows from this seed over \(\mathcal{G}_0\). At each later step, the current node receives a small set of earlier node outputs as prompt memory and generates a new text. We use \(t\) for the current generation step and \(m\) for a previous event. The generated history before step \(t\) is
\begin{equation*}
\mathcal{H}_t=\{e_m=(\tau_m,n_m,x_m):0\le m<t\},
\end{equation*}
where \(\tau_m\) is the timestamp, \(n_m\in\mathcal{N}\) is the node, and \(x_m\) is the text of event \(m\). Let \(k\ge 1\) be the maximum number of node representatives allowed in the prompt. A memory policy \(\pi\) maps the history, next timestamp, and next node to compact weighted memory, \(\mathcal{M}_t=\pi(\mathcal{H}_t,\tau_t,n_t)\). Each memory item keeps one representative from one node. If node \(j\) is retained for step \(t\), then \(r_t(j)\) denotes its latest previous generated event before \(\tau_t\). The memory has the form
\begin{equation*}
\mathcal{M}_t\subseteq
\{(j,r_t(j),w_{j,t}): j\in\mathcal{N},\ r_t(j)<t,\ w_{j,t}>0\}.
\end{equation*}
Here \(w_{j,t}\) is the weight assigned to node \(j\), and \(|\mathcal{M}_t|\le k\). When \(\mathcal{M}_t\neq\emptyset\), the selected weights satisfy
\begin{equation*}
\sum_{(j,r_t(j),w_{j,t})\in\mathcal{M}_t}w_{j,t}=1.
\end{equation*}
Memory selection means choosing which node representatives, and which weights, enter the next prompt.

The text transition is then explicit. Let \(a_i\) denote the fixed instruction string for node \(i\). Given \(\tau_t\), \(n_t\), and \(\mathcal{M}_t\), the prompt is
\begin{equation*}
\begin{aligned}
p_t=\operatorname{Prompt}\!\big(&
(\tau_t,n_t),\ a_{n_t},\\
&\{(w_{j,t},j,x_{r_t(j)}):(j,r_t(j),w_{j,t})\in\mathcal{M}_t\}\big).
\end{aligned}
\end{equation*}

Here \(\operatorname{Prompt}(\cdot)\) denotes deterministic formatting. In our implementation, the prompt states the target node, gives a short node-style instruction, and lists the selected predecessor texts with their node labels and normalized Hawkes weights. The weights are written as text annotations, so they guide the model through the prompt rather than acting as numerical controls. Appendix~\ref{app:prompt-example} gives a representative instantiated prompt.

\begin{figure*}[!t]
\centering
\includegraphics[width=0.86\textwidth]{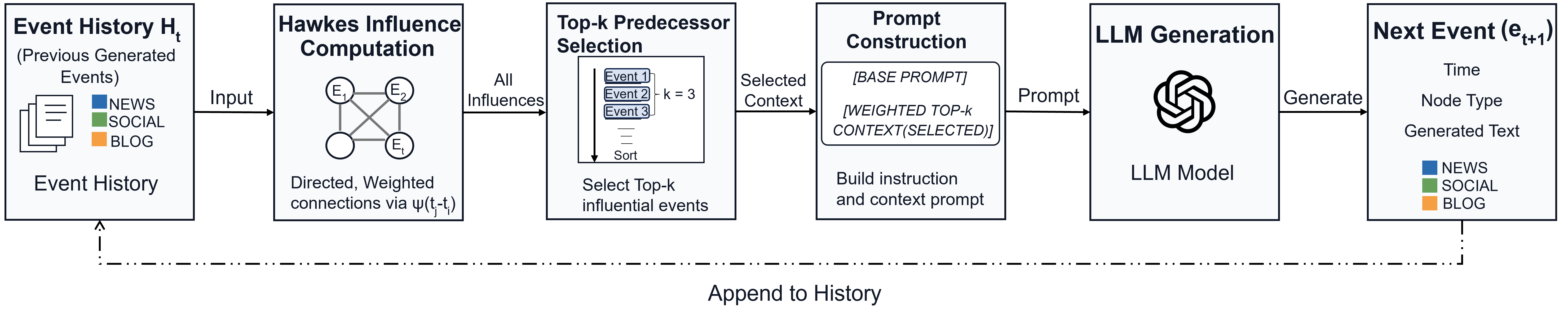}
\caption{Sequential agentic uncertainty loop in \HawkLLM. The generated history \(\mathcal{H}_t\) is the agent state; the Hawkes process selects weighted prompt memory \(\mathcal{M}_t\), the LLM generates \(x_t\), and the completed event \(e_t\) is appended to the history. Semantic alignment and local/global drift track trajectory-level uncertainty.}
\label{fig:framework}
\end{figure*}

When \(\mathcal{M}_t=\emptyset\), the prompt contains only the current event information and node instruction. Let \(g_{\mathrm{LLM}}\) denote the text generator. The next text is generated as
\begin{equation*}
x_t \sim g_{\mathrm{LLM}}(\cdot\mid p_t),
\end{equation*}
or deterministically as \(x_t=g_{\mathrm{LLM}}(p_t)\) when the decoding seed is fixed. The semantic propagation update appends the completed event:
\begin{equation*}
e_t=(\tau_t,n_t,x_t),\qquad \mathcal{H}_{t+1}=\mathcal{H}_t\cup\{e_t\}.
\end{equation*}
Only the text \(x_t\) is produced by the language model at step \(t\). The timestamp, node, and memory \(\mathcal{M}_t\) determine the prompt that conditions it. Figure~\ref{fig:framework} summarizes this loop.

\subsection{Multivariate Hawkes Point Process}
\label{sec:hawkes-process}
We use a multivariate Hawkes point process as the temporal influence model \citep{hawkes1971spectra,ogata1981lewis,rizoiu2017tutorial}. It estimates when each node is likely to activate and how much past activity from one node raises the future rate of another. For fitting, let \(\mathcal{D}=\{(\tau_m,n_m)\}_{m=1}^M\) denote an observed node-time event stream over horizon \([0,T]\). Here \(M\) is the number of observed events. Event \(m\) occurs at time \(\tau_m\in[0,T]\) and belongs to node \(n_m\in\mathcal{N}\). In this subsection, \(i\) indexes the node whose rate is being evaluated, and \(j\) indexes a node with a previous event. At continuous time \(s\), let \(\lambda_i(s)\) denote the conditional event rate for node \(i\). Let \(\mu_i\) denote its background rate. For each edge \((j,i)\in\mathcal{E}\), let \(\phi_{j,i}(u)\) denote the excitation kernel at lag \(u>0\). It describes how much a past event from node \(j\) raises the future rate of node \(i\). The event stream is modeled as
\begin{equation*}
\lambda_i(s)=\mu_i+\sum_{j:(j,i)\in\mathcal{E}} \sum_{\tau_m<s,\, n_m=j} \phi_{j,i}(s-\tau_m).
\end{equation*}
Our main model uses an exponential kernel. Here \(\alpha_{j,i}\ge 0\) is the directed excitation strength, and \(\beta>0\) is the decay rate:
\begin{equation*}
\phi_{j,i}(u)=\alpha_{j,i} e^{-\beta u}.
\end{equation*}
Larger \(\beta\) makes influence fade faster. For the exponential kernel, the integrated excitation matrix \(\mathbf{G}\) has entries \(G_{j,i}=\int_0^\infty \phi_{j,i}(u)\,du=\alpha_{j,i}/\beta\) for \((j,i)\in\mathcal{E}\), and \(G_{j,i}=0\) otherwise. Stability is assessed through the spectral radius \(\rho(\mathbf{G})\).

The weighted influence graph is \(\mathcal{G}=(\mathcal{N},\mathcal{E},\mathbf{G})\). The unweighted graph \(\mathcal{G}_0\) determines which influences are allowed, while \(G_{j,i}\) gives the fitted cumulative excitation from node \(j\) to node \(i\). Later, when building prompt memory, \HawkLLM turns this fitted influence into node-wise scores by applying temporal decay to earlier generated events from each node.

For a fixed decay value \(\beta\), we estimate \((\boldsymbol{\mu},\boldsymbol{\alpha})\). Here \(\boldsymbol{\mu}=(\mu_i)_{i\in\mathcal{N}}\) and \(\boldsymbol{\alpha}=(\alpha_{j,i})_{(j,i)\in\mathcal{E}}\). Let \(\lambda_i(s;\boldsymbol{\mu},\boldsymbol{\alpha},\beta)\) denote the corresponding conditional intensity. The log-likelihood term is
\begin{equation*}
\begin{aligned}
\ell_{\beta}(\boldsymbol{\mu},\boldsymbol{\alpha};\mathcal{D})
&= \sum_{m=1}^M \log \lambda_{n_m}(\tau_m;\boldsymbol{\mu},\boldsymbol{\alpha},\beta) \\
&\quad - \sum_{i\in\mathcal{N}} \int_0^T \lambda_i(s;\boldsymbol{\mu},\boldsymbol{\alpha},\beta)\,ds .
\end{aligned}
\end{equation*}
The fitted parameters solve
\begin{equation*}
(\hat{\boldsymbol{\mu}},\hat{\boldsymbol{\alpha}})
\in
\arg\max_{\boldsymbol{\mu}\ge 0,\,\boldsymbol{\alpha}\ge 0}
\left\{
\ell_{\beta}(\boldsymbol{\mu},\boldsymbol{\alpha};\mathcal{D})
-\eta\Omega(\boldsymbol{\alpha})
\right\}.
\end{equation*}
The term \(\Omega(\boldsymbol{\alpha})\) is a shrinkage penalty on excitation strengths, and \(\eta\ge 0\) controls the penalty strength. We select among fixed decay values by likelihood subject to stability and use the best stable exponential fit in the held-out simulation. Appendix~\ref{app:hawkes} reports fitting details.

\subsection{Hawkes-Based Semantic Propagation}
\label{sec:hawkes-propagation}
We now combine the Hawkes model with the text-generation loop. Let \(e_0=(\tau_0,n_0,x_0)\) be the seed event. The algorithm then produces \(L\) non-seed events \(e_1,\ldots,e_L\).

Before step \(t\), the history \(\mathcal{H}_t=\{e_m=(\tau_m,n_m,x_m):0\le m<t\}\) contains the seed and all earlier generated events. Its node-time projection is \(\mathcal{T}_t=\{(\tau_m,n_m):e_m\in\mathcal{H}_t\}\). Given the fitted Hawkes parameters \((\hat{\boldsymbol{\mu}},\hat{\boldsymbol{\alpha}},\hat{\beta})\), the conditional intensity for node \(i\) at candidate time \(r\) is evaluated from \(\mathcal{T}_t\) and written as \(\lambda_i(r\mid\mathcal{T}_t)\). At each step, the Hawkes process samples \(\tau_t\) and \(n_t\), constructs \(\mathcal{M}_t\), and passes the resulting prompt to the LLM.

Algorithm~\ref{alg:hawkes-step} gives the full \HawkLLM simulation loop. Event sampling produces the next timestamp and node, the memory policy produces \(\mathcal{M}_t\), and the final lines generate \(x_t\) and append \(e_t\) to the history.

\begin{algorithm}[tb]
\caption{\HawkLLM semantic propagation loop}
\label{alg:hawkes-step}
\footnotesize
\begin{algorithmic}[1]
\STATE {\bfseries Input:} seed \(e_0=(\tau_0,n_0,x_0)\), graph \(\mathcal{G}_0=(\mathcal{N},\mathcal{E})\), budget \(k\), length \(L\), fitted Hawkes model \((\hat{\boldsymbol{\mu}},\hat{\boldsymbol{\alpha}},\hat{\beta})\), thresholds \((\epsilon_{\mathrm{raw}},\epsilon_{\mathrm{norm}})\), node instructions \(\{a_i:i\in\mathcal{N}\}\), text generator \(g_{\mathrm{LLM}}\)
\STATE {\bfseries Output:} simulated history \(\mathcal{H}_{L+1}\)
\STATE \(\mathcal{H}_1\gets\{e_0\}\)
\FOR{\(t=1,\ldots,L\)}
\STATE \(\mathcal{T}_t\gets\{(\tau_m,n_m):e_m\in\mathcal{H}_t\}\)
\STATE \(s\gets\tau_{t-1}\)
\STATE {\bfseries Event sampling}
\STATE \(\Lambda_t(r)\gets\sum_{i\in\mathcal{N}}\lambda_i(r\mid\mathcal{T}_t)\)
\STATE sample \(\tau_t\) by thinning from time \(s\) using \(\Lambda_t\)
\STATE sample \(n_t=i\) with probability \(\lambda_i(\tau_t\mid\mathcal{T}_t)/\Lambda_t(\tau_t)\)
\STATE {\bfseries Memory policy}
\STATE \(\mathcal{J}_t\gets\{j\in\mathcal{N}:(j,n_t)\in\mathcal{E},\ \exists m<t\text{ with }n_m=j,\ \tau_m<\tau_t\}\)
\STATE \(r_t(j)\gets\max\{m<t:n_m=j,\ \tau_m<\tau_t\}\) for \(j\in\mathcal{J}_t\)
\STATE \(h_{j,t}\gets\sum_{m<t:\ n_m=j,\ \tau_m<\tau_t}\exp[-\hat{\beta}(\tau_t-\tau_m)]\) for \(j\in\mathcal{J}_t\)
\STATE \(q_{j,t}\gets\hat{\alpha}_{j,n_t}h_{j,t}\) for \(j\in\mathcal{J}_t\)
\STATE \(Q_t\gets\sum_{\ell\in\mathcal{J}_t}q_{\ell,t}\)
\IF{\(Q_t=0\)}
\STATE \(\mathcal{I}_t\gets\emptyset\)
\ELSE
\STATE \(\bar q_{j,t}\gets q_{j,t}/Q_t\) for \(j\in\mathcal{J}_t\)
\STATE \(\widetilde{\mathcal{J}}_t\gets\{j\in\mathcal{J}_t:q_{j,t}\ge\epsilon_{\mathrm{raw}},\ \bar q_{j,t}\ge\epsilon_{\mathrm{norm}}\}\)
\STATE \(\mathcal{I}_t\gets\operatorname{TopK}_k(\widetilde{\mathcal{J}}_t;\ q_{j,t})\)
\ENDIF
\IF{\(\mathcal{I}_t=\emptyset\)}
\STATE \(\mathcal{M}_t\gets\emptyset\)
\ELSE
\STATE \(w_{j,t}\gets q_{j,t}/\sum_{\ell\in\mathcal{I}_t}q_{\ell,t}\) for \(j\in\mathcal{I}_t\)
\STATE \(\mathcal{M}_t\gets\{(j,r_t(j),w_{j,t}):j\in\mathcal{I}_t\}\)
\ENDIF
\STATE {\bfseries Text update}
\STATE \(p_t\gets\operatorname{Prompt}((\tau_t,n_t),a_{n_t},\{(w_{j,t},j,x_{r_t(j)}):(j,r_t(j),w_{j,t})\in\mathcal{M}_t\})\)
\STATE sample \(x_t\sim g_{\mathrm{LLM}}(\cdot\mid p_t)\)
\STATE \(e_t\gets(\tau_t,n_t,x_t)\), \(\mathcal{H}_{t+1}\gets\mathcal{H}_t\cup\{e_t\}\)
\ENDFOR
\STATE \textbf{return} \(\mathcal{H}_{L+1}\)
\end{algorithmic}
\end{algorithm}

\subsubsection{Event Simulation by Thinning}
The event-sampling block of Algorithm~\ref{alg:hawkes-step} samples the next timestamp and node from the fitted multivariate Hawkes process using Ogata-style thinning \citep{ogata1981lewis}. It first extracts \(\mathcal{T}_t\), the event-time and node projection of \(\mathcal{H}_t\). At \(t=1\), this projection contains only the seed time and seed node. The total intensity is \(\Lambda_t(s)=\sum_{i\in\mathcal{N}}\lambda_i(s\mid\mathcal{T}_t)\). Starting from the previous event time, thinning proposes a waiting time from an exponential distribution with rate \(\bar{\Lambda}\), where \(\bar{\Lambda}\) is a local upper bound on \(\Lambda_t\). A proposal time \(\tilde{s}\) is accepted with probability \(\Lambda_t(\tilde{s})/\bar{\Lambda}\). After acceptance, we set \(\tau_t=\tilde{s}\) and sample the node \(n_t=i\) with probability proportional to \(\lambda_i(\tau_t\mid\mathcal{T}_t)\).

This sampling step uses only event times and nodes. The history texts \(\{x_m:e_m\in\mathcal{H}_t\}\) enter only after \(\tau_t\) and \(n_t\) are sampled, when the memory policy chooses predecessor texts for the LLM prompt.

\subsubsection{Hawkes Memory Policy}
The memory-policy block of Algorithm~\ref{alg:hawkes-step} constructs \(\mathcal{M}_t\) for the sampled timestamp and node. We aggregate Hawkes contributions at the node level. The Hawkes state still uses all earlier event times through exponential decay, but the text prompt contains at most one representative text per retained node: the latest generated text from that node.

For each eligible node \(j\), define
\begin{equation*}
r_t(j)=\max\{m<t:n_m=j,\ \tau_m<\tau_t\},
\end{equation*}
when such an event exists. The node-wise decayed state is
\begin{equation*}
h_{j,t}=
\sum_{m<t:\ n_m=j,\ \tau_m<\tau_t}
\exp[-\hat{\beta}(\tau_t-\tau_m)].
\end{equation*}
The node-wise Hawkes contribution toward the current node \(n_t\) is then
\begin{equation*}
q_{j,t}=\hat{\alpha}_{j,n_t}h_{j,t}.
\end{equation*}
The score is larger when node \(j\) has stronger learned excitation toward the current node and when its recent events have not decayed away.

Let
\begin{equation*}
\mathcal{J}_t=\{j\in\mathcal{N}:(j,n_t)\in\mathcal{E},\ r_t(j)\text{ is defined}\}.
\end{equation*}
Let \(Q_t=\sum_{\ell\in\mathcal{J}_t}q_{\ell,t}\). If \(Q_t=0\), the selected set is empty. Otherwise, define the normalized contribution \(\bar q_{j,t}=q_{j,t}/Q_t\). The policy removes negligible nodes using a raw-score threshold \(\epsilon_{\mathrm{raw}}\) and a normalized-contribution threshold \(\epsilon_{\mathrm{norm}}\):
\begin{equation*}
\widetilde{\mathcal{J}}_t
=
\left\{
j\in\mathcal{J}_t:
q_{j,t}\ge \epsilon_{\mathrm{raw}},
\ \bar q_{j,t}\ge \epsilon_{\mathrm{norm}}
\right\}.
\end{equation*}
It then keeps at most \(k\) nodes with the largest remaining scores:
\begin{equation*}
\mathcal{I}_t
=
\operatorname{TopK}_k(\widetilde{\mathcal{J}}_t;\ q_{j,t}).
\end{equation*}
When \(\mathcal{I}_t\neq\emptyset\), the prompt-memory weights are the selected scores normalized to sum to one:
\begin{equation*}
w_{j,t}=\frac{q_{j,t}}{\sum_{\ell\in\mathcal{I}_t}q_{\ell,t}},\qquad j\in\mathcal{I}_t.
\end{equation*}
The Hawkes memory policy returns
\begin{equation*}
\mathcal{M}_t=\{(j,r_t(j),w_{j,t}):j\in\mathcal{I}_t\}.
\end{equation*}
When \(\mathcal{I}_t=\emptyset\), we set \(\mathcal{M}_t=\emptyset\), and the prompt uses only the current event information and the style instruction for the current node. The top-\(k\) step is an engineering constraint for compact prompts; the Hawkes process itself remains the temporal influence model. The filtering thresholds are implementation parameters.

The text-update block then assembles \(p_t\), samples \(x_t\), and appends \(e_t\). Thus the Hawkes process controls event timing, current node, and selected memory, while the language model verbalizes the next event.

\section{Experimental Setting}
\subsection{Data and Event Stream}
We use Global Database of Events, Language, and Tone (GDELT) article metadata \citep{leetaru2013gdelt}. We select a recent event window for \textit{Artemis II} coverage using the query terms \texttt{"Artemis II"} and \texttt{"Artemis 2"}. The working window spans April 1--11, 2026 UTC. GDELT timeline metadata indicate a rough scale of 10{,}525 matching articles over the window. The modeling dataset uses a capped sample of 250 article records. After URL/title deduplication, the final event stream contains 248 English-language events over approximately 263 hours. Tied timestamps are broken by small within-group offsets so that point-process fitting and chronological train/test splitting are well defined. The retained fields are timestamp, article domain, node, language, title, and URL.

Nodes are defined as hand-curated outlet categories, not as semantic topic labels. Table~\ref{tab:node-map} reports the final five-node grouping.
\begin{table}[!ht]
\centering
\small
\setlength{\tabcolsep}{5pt}
\caption{Node categories and event counts in the modeling dataset.}
\label{tab:node-map}
\begin{tabular}{lr}
\toprule
Node & Count \\
\midrule
\texttt{local\_tv} & 69 \\
\texttt{mass\_market} & 37 \\
\texttt{specialist\_science\_tech} & 25 \\
\texttt{business\_finance} & 31 \\
\texttt{general\_news} & 86 \\
\bottomrule
\end{tabular}
\end{table}
This grouping is a hand-curated, task-specific media taxonomy for the Artemis II window. Representative domains include local broadcast affiliates for \texttt{local\_tv}, tabloid or regional commercial outlets for \texttt{mass\_market}, science and space outlets for \texttt{specialist\_science\_tech}, finance outlets for \texttt{business\_finance}, and a residual \texttt{general\_news} category for the remaining domains.

\paragraph{Generation setup.}
The text generator is Qwen2.5 run through a local Ollama backend, \texttt{ollama:qwen2.5:latest} \citep{bai2024qwen25}, decoded with temperature \(0.35\), top-p \(0.9\), and at most 75 new tokens. This generator is shared across node agents; the node label and node instruction determine the role being simulated at each step. Appendix~\ref{app:gen-settings} gives the remaining generation settings.

\paragraph{Seed event.}
Seed choice depends on the experimental setting. For the held-out train/test evaluation reported in Tables~\ref{tab:main}--\ref{tab:ksens_main}, each post-split simulation starts from the last training event and is evaluated over the subsequent test window, so no test-set title is used to initialize generation. For the full-data illustrative simulation and the qualitative example in Table~\ref{tab:qualitative}, we use the earliest observed Artemis II title as the seed:
\par\smallskip
\noindent\emph{Moon rocket and weather are on NASA side for the first astronaut launch in decades.}
\par\smallskip
In both settings, the seed supplies \(e_0=(\tau_0,n_0,x_0)\) and anchors the generated trajectory before the Hawkes process samples subsequent timestamp/node pairs.

\subsection{Baselines}
We compare \HawkLLM memory selection against two simple heuristic baselines. Both reuse the same event sequence, prompt format, text generator, and evaluation pipeline, changing only the predecessor-selection rule:
\begin{itemize}[leftmargin=1.2em]
    \item \textbf{Chronological last-$k$}: the \(k\) most recent past generated events with uniform weights.
    \item \textbf{Random-$k$}: \(k\) uniformly sampled past generated events with uniform weights.
\end{itemize}
Unless otherwise stated, all methods use \(k=3\).

\subsection{Train-Test Split and Matching}
\label{sec:eval-protocol}
For the held-out evaluation, we split the 248-event dataset chronologically into train (198 events) and test (50 events). The Hawkes process is refit on train only using a stable exponential specification. Simulation is then constrained to the held-out test horizon. Appendix~\ref{app:eval} gives split counts and matching details.

For each generated non-seed event, we find same-node real test events within \(\pm 12\) hours. If none exist, we relax to \(\pm 24\) hours. In total, 62 generated non-seed events are evaluated across three post-split runs. All 62 find a match, with 58 primary-window matches and 4 relaxed-window matches.

\subsection{Evaluation Diagnostics}
\label{sec:diagnostics}
We reserve held-out texts for evaluation. Newsroom choice, outlet access, framing, and repetition across outlets all affect which article is actually written at a given time, so we evaluate local semantic agreement rather than exact continuation prediction. For a generated event \(e_t=(\tau_t,n_t,x_t)\), let \(\mathcal{R}_t\) be the matched set of real test-set titles from the same local region of the cascade. In this GDELT case study, those references are held-out Artemis II article titles.

When \(\mathcal{R}_t\neq\emptyset\), we embed the generated text and reference texts with the evaluation embedding function \(\mathbf{z}(\cdot)\). In the reported experiments, \(\mathbf{z}\) is computed with the same local Ollama backend used by the saved real-vs-similarity and drift outputs, \texttt{ollama:qwen2.5:latest} \citep{bai2024qwen25}. This makes the evaluation reproducible in the local pipeline, while keeping the metric at the level of a semantic-neighborhood diagnostic; independent embedding backends, human judgments, and factuality checks are natural extensions. Semantic alignment is the cosine similarity between the generated-text embedding and the average reference embedding:
\begin{equation*}
S_t = \cos\!\left(\mathbf{z}(x_t), \frac{1}{|\mathcal{R}_t|}\sum_{r \in \mathcal{R}_t} \mathbf{z}(r)\right),
\end{equation*}
where higher values indicate closer agreement with the local held-out set. If \(|\mathcal{R}_t|=1\), this is simply cosine similarity between \(x_t\) and the single matched reference text. We report mean \(S_t\), temporal trend, and late-stage behavior, summarized by averaging \(S_t\) over the final 20\% of matched simulated events within each run.

We also decompose uncertainty into global and local drift. Let \(x_0\) be the seed text. When \(\mathcal{M}_t\neq\emptyset\), let
\begin{equation*}
\bar{\mathbf{z}}_t=\sum_{(j,r_t(j),w_{j,t})\in\mathcal{M}_t} w_{j,t}\mathbf{z}(x_{r_t(j)})
\end{equation*}
be the weighted predecessor centroid from the memory policy. We define
\begin{align*}
D_t^{\text{global}} &= 1-\cos(\mathbf{z}(x_t),\mathbf{z}(x_0)), \\
D_t^{\text{local}} &= 1-\cos(\mathbf{z}(x_t),\bar{\mathbf{z}}_t).
\end{align*}
Global drift measures distance from the seed text. Local drift measures distance from the weighted predecessor memory that directly conditioned the current step. Local drift is undefined when no valid predecessor prompt memory exists.

\section{Results}

\subsection{Semantic Alignment Over the Generated Trajectory}

Table~\ref{tab:main} reports held-out semantic alignment under the matched compact prompt-memory budget (\(k=3\)). Late-stage \(S_t\) is the average over the final 20\% of matched simulated events, as defined in Section~\ref{sec:diagnostics}. In this setting, \HawkLLM has the highest mean and late-stage alignment, with the largest separation near the tail of the generated trajectory.

\begin{table}[!ht]
\centering
\small
\setlength{\tabcolsep}{3pt}
\caption{Held-out semantic alignment \(S_t\) under the matched compact prompt-memory budget \(k=3\). Higher values are better.}
\label{tab:main}
\begin{tabular}{lccc}
\toprule
Method & Mean \(S_t\) & Trend & Late-stage \(S_t\) \\
\midrule
HawkesLLM & \textbf{0.636} & increasing & \textbf{0.682} \\
Chronological last-$k$ & 0.581 & decreasing & 0.541 \\
Random-$k$ & 0.621 & decreasing & 0.594 \\
\bottomrule
\end{tabular}
\end{table}

\begin{figure*}[!t]
\centering
\includegraphics[width=\linewidth]{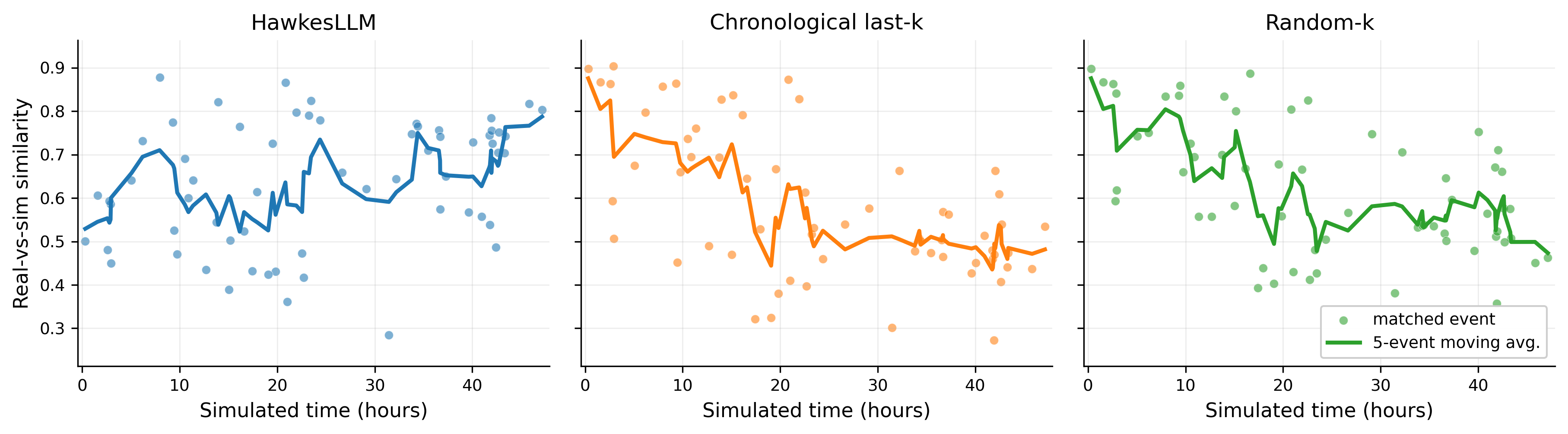}
\caption{Sequential semantic alignment \(S_t\) over simulated time for \HawkLLM, chronological last-$k$, and random-$k$ on the held-out test window. Panels show matched generated events and a 5-event moving average; higher values indicate closer alignment to the local held-out reference set. The seed event comes from the training window, so the plotted curves begin with matched generated events in the test window.}
\label{fig:similarity}
\end{figure*}

The trend column is computed directly from \(S_t\) over simulated time. \HawkLLM is the only method in this comparison whose semantic alignment increases over the generated trajectory. Chronological and random memory selection decrease. Figure~\ref{fig:similarity} shows the corresponding trajectory-level semantic-alignment curves.

These comparisons are descriptive case-study evidence from a limited held-out sample with dependent generated events. They should be read as diagnostics for this GDELT window, not as a broad benchmark claim.

\subsection{Effect of Prompt-Memory Budget}

The parameter \(k\) controls prompt-memory size. For \HawkLLM, it truncates a weighted influence structure; for the baselines, it directly sets how many predecessor events are provided. Table~\ref{tab:ksens_main} reports the \(k\)-sensitivity comparison using semantic alignment \(S_t\).

\begin{table}[!ht]
\centering
\small
\setlength{\tabcolsep}{4pt}
\caption{Prompt-memory budget sensitivity using semantic alignment \(S_t\) in a representative run. Higher is better.}
\label{tab:ksens_main}
\begin{tabular}{lcccc}
\toprule
Method & \(k\) & Mean \(S_t\) & \(S_t\) trend & Late-stage \(S_t\) \\
\midrule
HawkesLLM & 3 & 0.635 & increasing & \textbf{0.703} \\
HawkesLLM & 5 & 0.634 & increasing & \textbf{0.703} \\
HawkesLLM & 7 & 0.634 & increasing & \textbf{0.703} \\
Chronological & 3 & 0.578 & decreasing & 0.497 \\
Chronological & 5 & 0.556 & decreasing & 0.454 \\
Chronological & 7 & \textbf{0.694} & flat & 0.636 \\
Random & 3 & 0.633 & decreasing & 0.557 \\
Random & 5 & 0.597 & decreasing & 0.537 \\
Random & 7 & 0.642 & decreasing & 0.627 \\
\bottomrule
\end{tabular}
\end{table}

\HawkLLM changes little as \(k\) increases because most events use fewer than three meaningful weighted neighbors; extra slots rarely add useful prompt memory.

The heuristic baselines can benefit from larger prompt memory. Chronological last-$k$ at \(k=7\) achieves higher mean semantic alignment than \HawkLLM in this representative run, but its advantage fades later in the trajectory; its late-stage \(S_t\) remains below that of \HawkLLM. The takeaway is narrower but useful: \HawkLLM is strongest when prompt memory is compact and late-stage behavior matters. A broader comparison should aggregate this \(k\)-sensitivity over repeated runs and match methods by token budget as well as by item count.

\subsection{Local and Global Components of Sequential Uncertainty}

Across repeated simulations, the same drift pattern appears. In the post-split setting, the average mean global drift is \(0.450 \pm 0.019\), while the average mean local drift is \(0.185 \pm 0.072\). Global drift exceeds local drift in all runs.

\begin{figure}[!ht]
\centering
\includegraphics[width=\linewidth]{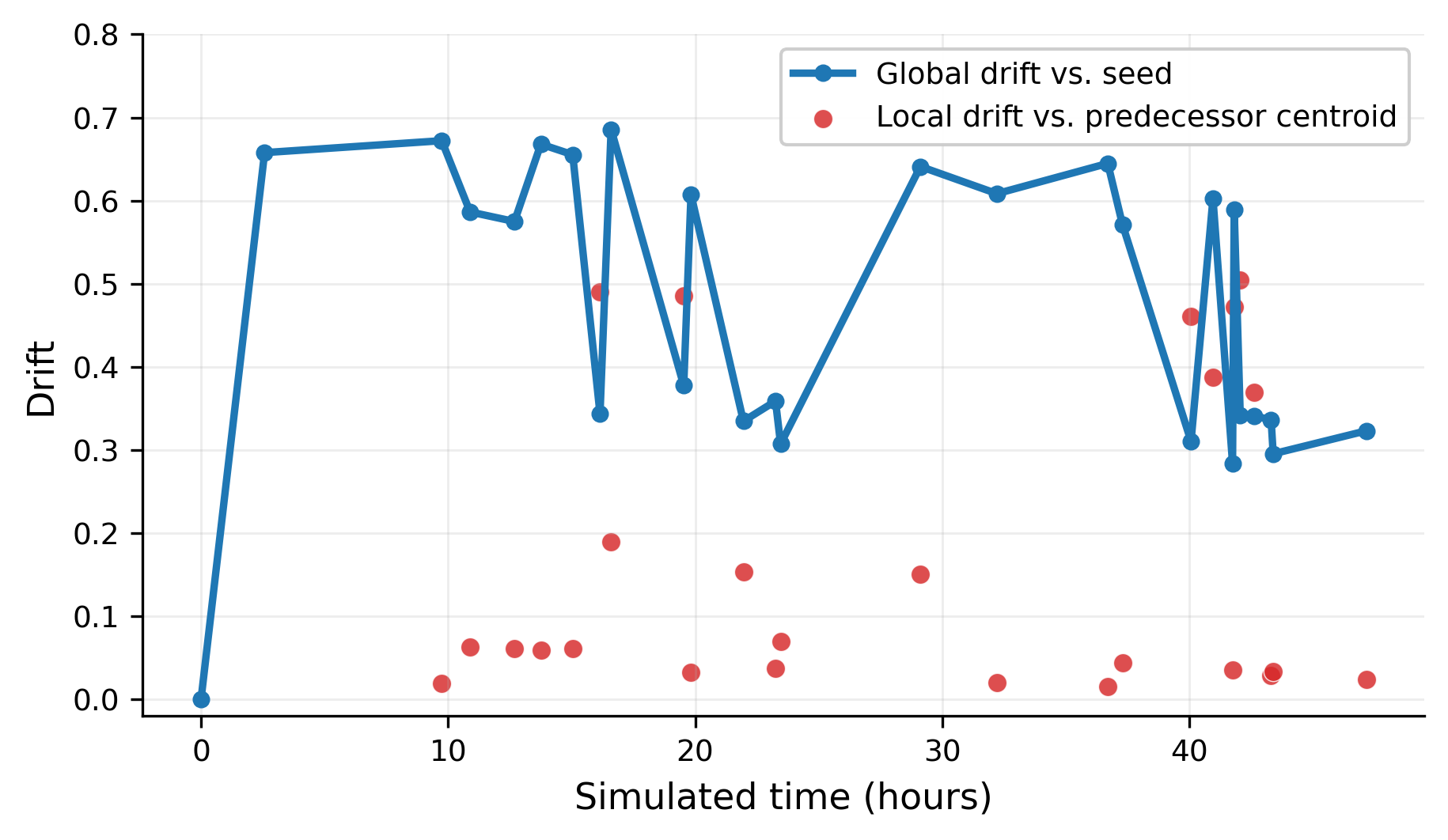}
\caption{Global and local drift over a representative \HawkLLM run. Global drift is measured relative to the seed text, while local drift is measured relative to the weighted prompt memory.}
\label{fig:drift}
\end{figure}

These runs show generated cascades accumulating global uncertainty while staying locally stable: the trajectory can remain close to its immediate weighted prompt memory and still move away from the seed over longer horizons. Individual runs need not be monotone; the repeated post-split runs consistently show global drift above local drift. Figure~\ref{fig:drift} illustrates this separation over a representative \HawkLLM run.

\begin{figure}[!ht]
\centering
\includegraphics[width=\linewidth,trim={0 0 0 18bp},clip]{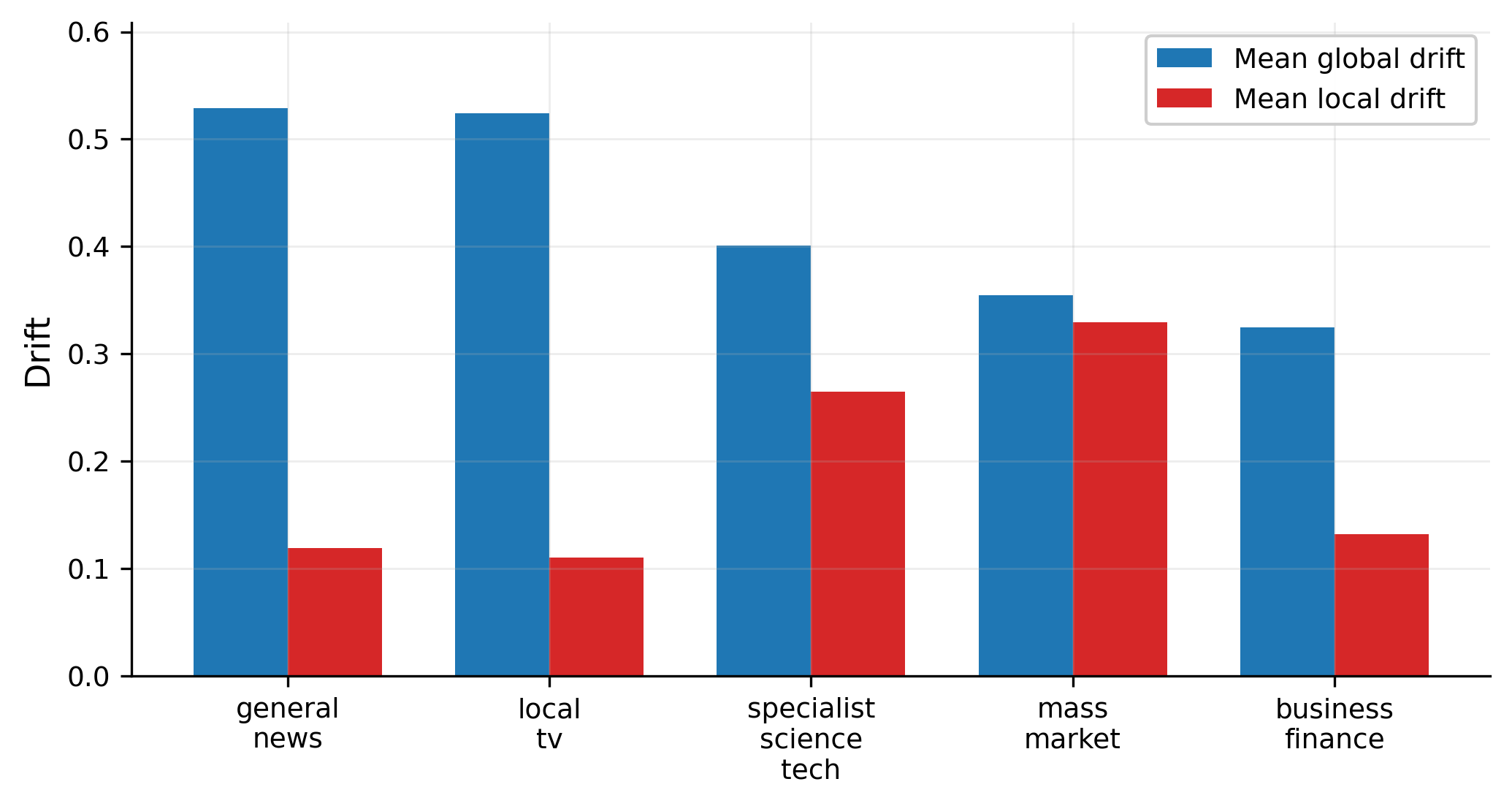}
\caption{Node-conditioned mean global and local drift across \HawkLLM simulations. Drift varies by node.}
\label{fig:node-drift}
\end{figure}

Figure~\ref{fig:node-drift} breaks the drift diagnostic down by node. The global/local separation is visible across most node categories.

\subsection{Qualitative Propagation Example}

Table~\ref{tab:qualitative} shows an illustrative trajectory generated by \HawkLLM. 
Rather than listing every event, we select representative events that show how the generated cascade changes framing across source types. 
The trajectory begins with a general-news seed about favorable launch conditions, then shifts toward technical mission-status language, business implications, local community framing, and eventually more public-facing launch narratives. 
This example is qualitative only, but it illustrates the behavior measured by the quantitative diagnostics: the sequence stays within the Artemis II semantic region while its emphasis changes across nodes and over time.

\begin{table}[!ht]
\centering
\scriptsize
\setlength{\tabcolsep}{3pt}
\caption{Illustrative \HawkLLM-generated cascade. We show selected representative events from one run to highlight how source type changes the framing of the same evolving Artemis II topic.}
\label{tab:qualitative}
\begin{tabular*}{\linewidth}{@{\extracolsep{\fill}} r l p{0.62\linewidth}@{}}
\toprule
Time & Node & Generated text \\
\midrule
0.0 & general & Moon rocket and weather are on NASA side for the first astronaut launch in decades. \\
0.4 & general & NASA sets course for space with Artemis II, as favorable conditions align for historic crewed mission. \\
3.2 & sci./tech & Initial data from the Artemis II mission indicates stable performance across all systems, with ongoing analysis of trajectory adjustments and spacecraft health metrics. \\
8.2 & finance & Artemis II Mission Readies for Launch, Signaling Major Advancements in Space Exploration Ventures. \\
11.6 & mass & NASA readies its crew for the groundbreaking Artemis II mission, signaling a major leap in space exploration under favorable conditions. \\
13.1 & local TV & Local astronauts are getting ready for their big journey as NASA gears up for the Artemis II mission, marking an exciting step forward in space travel. \\
23.1 & mass & NASA's Artemis II mission gets set to blast off, signaling a huge leap forward in our quest to explore the cosmos under perfect conditions. \\
70.3 & sci./tech & Engineers at Kennedy Space Center are finalizing pre-launch procedures for Artemis II, heralding a significant step forward in NASA's lunar exploration mission. \\
\bottomrule
\end{tabular*}
\end{table}

\section{Conclusion}
This paper introduced \HawkLLM, a framework for monitoring semantic uncertainty propagation in iterative text-simulation systems. The Hawkes process exposes node-wise influence scores that determine which generated memories enter future prompts, so uncertainty can be tracked over the full trajectory. In the GDELT case study, \HawkLLM yields stronger late-stage semantic alignment under a compact prompt-memory budget. Local/global drift diagnostics show how uncertainty can accumulate globally even when individual steps remain locally stable.

The gain is specific rather than sweeping. The Qwen generator already has a strong prior for plausible Artemis II text; \HawkLLM adds structure around which node memories enter the prompt. Its role is to make node-wise information flow and uncertainty monitoring explicit around the generator. Although we study news-style event propagation, the same formulation can be used for social-media narratives, multi-step agent interactions, and iterative workflows where earlier generated text becomes later context.

The study is diagnostic. It uses a sampled GDELT article list, title-level text, a hand-curated node taxonomy, and a held-out test set whose \texttt{specialist\_science\_tech} coverage is sparse. The generator can also introduce artifacts, including occasional mixed-language outputs. Because the reported embedding metric uses the same Qwen/Ollama model family as the generator, the next evaluation step is to pair these diagnostics with independent embedding backends, human judgments, and factuality checks. Richer event texts, additional domains, and calibrated threshold rules would help determine when semantic monitoring is reliable enough for operational use.

\section*{Impact Statement}
This paper presents a research baseline for studying semantic uncertainty in multi-step text simulation. The same machinery that makes plausible news-style trajectories useful for analysis also creates misuse risks. We use the framework here as a diagnostic research instrument; deployment would require content safeguards, provenance controls, and separate reliability validation.

\bibliography{refs}
\bibliographystyle{icml2026}

\appendix

\section{Instantiated Prompt Example}
\label{app:prompt-example}

The prompt format is fixed across methods. \HawkLLM changes only which node-labeled predecessor texts and weights enter the memory block. The example below shows one instantiated prompt. The weights appear as text annotations, so they provide relative-importance cues rather than hard numerical controls inside the language model.

\begin{promptbox}{Instantiated Example}
\promptfield{Role}{You are simulating one item in a Hawkes-driven cross-node news cascade.}
\promptfield{Task}{Write exactly one concise English sentence or headline-like update.}
\promptfield{Constraints}{Use only the weighted predecessor texts below; do not mention weights, simulations, models, or prompts. Do not copy predecessor wording verbatim. Preserve the core Artemis II subject while allowing natural semantic drift.}
\promptfield{Target node}{\texttt{local\_tv}}
\promptfield{Node style}{Write like a local TV news web update: clear, public-facing, practical, and locally relatable.}
\promptfield{Simulated time since seed}{15.04 hours}
\promptfield{Weighted predecessor context}{}
\promptcontext{1}{0.81}{local\_tv}{Artemis II crew making great strides in their preparations for the mission, according to latest updates from NASA. Local teams are keeping a close eye on their progress and wish them all the best for this historic journey.}
\promptcontext{2}{0.19}{general\_news}{Crew aboard Artemis II continues to prepare for historic space mission, overcoming initial hurdles as NASA reports smooth progress.}
\promptfield{Output}{Only the generated news item.}
\end{promptbox}

\section{Hawkes Fitting Details}
\label{app:hawkes}
The full-data exponential grid included eight fixed decay values from 1/72 to 1/6 per hour. The best-overall exponential fit had \(\beta=0.1667\) and \(\rho(\mathbf{G})=1.0535\), so it was not used for simulation. The best stable exponential fit had \(\beta=0.0833\), \(\rho(\mathbf{G})=0.8708\), log-likelihood \(-556.821\), AIC \(1175.641\), and BIC \(1284.558\). The Gaussian-basis comparison used truncated normal bases centered at 6, 24, and 72 hours with standard deviations 4, 10, and 20 hours, respectively; it achieved \(\rho(\mathbf{G})=0.6836\) and log-likelihood \(-569.693\).

For the chronological 80/20 train/test evaluation, Hawkes was refit on train only. The resulting stable exponential model has \(\beta=0.1667\) and \(\rho(\mathbf{G})=0.8467\). That train-only refit was used for the held-out uncertainty evaluation.

\section{Generation Settings}
\label{app:gen-settings}
Table~\ref{tab:gen-settings} reports the decoding and prompt-memory settings used in the simulations.

\begin{table}[!ht]
\centering
\small
\setlength{\tabcolsep}{5pt}
\caption{Generation and prompt-memory settings.}
\label{tab:gen-settings}
\begin{tabular}{lr}
\toprule
Setting & Value \\
\midrule
Generator & Qwen2.5 via Ollama \\
Temperature & 0.35 \\
Top-p & 0.9 \\
Max new tokens & 75 \\
Main prompt budget \(k\) & 3 \\
Normalized threshold \(\epsilon_{\mathrm{norm}}\) & 0.03 \\
Raw threshold \(\epsilon_{\mathrm{raw}}\) & \(10^{-5}\) \\
Event cap & 40 \\
\bottomrule
\end{tabular}
\end{table}

The held-out simulations start from the last training event and then run over the test window. The full-data exploratory run used the earliest observed Artemis II title as its seed:
\begin{quote}
\small
Moon rocket and weather are on NASA side for the first astronaut launch in decades.
\end{quote}

\section{Held-Out Evaluation Details}
\label{app:eval}
The chronological 80/20 split yields 198 train events and 50 test events. Table~\ref{tab:split-counts} reports the per-node counts.

\begin{table}[!ht]
\centering
\small
\setlength{\tabcolsep}{5pt}
\caption{Per-node train/test counts under the chronological split.}
\label{tab:split-counts}
\begin{tabular}{lrr}
\toprule
Node & Train & Test \\
\midrule
\texttt{local\_tv} & 51 & 18 \\
\texttt{mass\_market} & 22 & 15 \\
\texttt{specialist\_science\_tech} & 22 & 3 \\
\texttt{business\_finance} & 26 & 5 \\
\texttt{general\_news} & 77 & 9 \\
\bottomrule
\end{tabular}
\end{table}

Across three post-split runs, 62 generated non-seed events were evaluated. All 62 found same-node matches, with 58 within \(\pm 12\) hours and 4 using the \(\pm 24\)-hour fallback. The similarity curves begin at the first matched generated test-window event, not at the training-side seed.

\paragraph{Additional drift summaries.}
In full-data repeated runs, the same global/local separation appears: average mean global drift is \(0.440 \pm 0.100\), and average mean local drift is \(0.153 \pm 0.049\).

\end{document}